\documentclass[runningheads]{llncs}
\usepackage{amsmath,amssymb,amsfonts}
\usepackage{mathtools}
\usepackage{algorithm}
\usepackage{algpseudocode}
\usepackage[colorlinks, linkcolor=blue, urlcolor=blue, citecolor=blue]{hyperref}
\usepackage[T1]{fontenc}
\usepackage{graphicx}
\usepackage{booktabs}
\usepackage[misc]{ifsym}
\usepackage{booktabs, paralist}
\newcommand{\corr}{(\Letter)}
\usepackage{mwe}

\begin{document}

\title{Offline Imitation of Badminton Player Behavior via Experiential Contexts and Brownian Motion}

\titlerunning{RallyNet: Offline Imitation of Badminton Player Behavior}

\author{Kuang-Da Wang\and
Wei-Yao Wang\and
Ping-Chun Hsieh\and
Wen-Chih Peng \corr}

\authorrunning{K. Wang et al.}

\institute{National Yang Ming Chiao Tung University, Hsinchu, Taiwan \email{gdwang.cs10@nycu.edu.tw, sf1638.cs05@nctu.edu.tw, pinghsieh@nycu.edu.tw, wcpeng@cs.nycu.edu.tw}}

\maketitle 

\begin{abstract}
In the dynamic and rapid tactic involvements of turn-based sports, badminton stands out as an intrinsic paradigm that requires alter-dependent decision-making of players.
While the advancement of learning from offline expert data in sequential decision-making has been witnessed in various domains (e.g., video games), how to imitate the behaviors of human players from rallies consisting of multiple players in offline badminton matches has remained underexplored.
Replicating opponents' behavior benefits players by allowing them to undergo strategic development with direction before matches.
However, directly applying existing methods suffers from the inherent hierarchy of the match and the compounding effect due to the turn-based nature of players alternatively taking actions.
In this paper, we propose RallyNet, a novel hierarchical offline imitation learning model for badminton player behaviors: 
(i) RallyNet captures players' decision dependencies by modeling decision-making processes as a contextual Markov decision process.
(ii) To facilitate the decision-making of an agent, RallyNet leverages the experience to generate \textit{context} as the agent's intent in the rally.
(iii) To generate more realistic behavior, RallyNet leverages Geometric Brownian Motion (GBM) to model the interactions between players by introducing a valuable inductive bias for learning player behaviors.
In this manner, RallyNet links player intents with interaction models with GBM, providing an understanding of real interactions for sports analytics.
We extensively validate RallyNet with the largest available real-world badminton dataset consisting of men's and women's singles, demonstrating its ability to adeptly imitate player behaviors. 
The results illustrate RallyNet's superiority, surpassing offline imitation learning methods and state-of-the-art turn-based approaches by at least 16\% in the mean of rule-based agent normalization score.
In addition, several practical use cases showcase the applicability of RallyNet. 

\end{abstract}

\section{Introduction}
The exploration of simulating agents' behaviors based on historical data has broad applicability across various domains.
Whether it involves replicating specific scenarios for sports analytics \cite{wang2021exploring}, autonomous driving \cite{bronstein2022hierarchical}, or robots \cite{singh2020scalable}, these scenarios can be effectively framed as simulations of agents' behaviors characterized by intricate interactions and decision-making processes.
In sports analytics, one of the major goals is to understand and investigate the tactics of teams and individuals, which can be achieved by analyzing behavioral records to reproduce strategic behaviors.
The applicability to the replication of player strategic behavior could assist in devising winning strategies \cite{le2017data} and sports broadcasting \cite{sport_broadcasting}. 

However, unlike basketball and football, where players dictate their own positions, a player's state (e.g., receiving position) is determined by the opponent in badminton, a typical turn-based sport.
As a result, conventional imitation learning methods employed in basketball \cite{basketball_bc} and football \cite{le2017data} cannot be directly applied to turn-based sports since finding authentic opponents for the agent to interact with and enhance its strategies in a realistic game scenario is impractical.
Therefore, we focus on leveraging offline behavioral records to replicate turn-based player behavior, employing offline imitation learning (IL) since it provides insights from past match data, enabling coaches and players to infer winning strategies. 
To effectively apply offline IL in badminton, three goals are defined as key criteria to assess the model's suitability for badminton scenarios:
\begin{compactitem}
    \item \textbf{Behavioral Sequence Similarity:} The learned agent's behavior sequences should closely mirror real-world rally content, capturing the order of shot types, shuttlecock trajectories, and player movement patterns.
    \item \textbf{Rally Duration Realism:} To depict how actual players participate in matches, it is crucial for the agents to reproduce the duration of rallies.
    \item \textbf{Outcome Consistency:} Interactions between learned agents should yield consistent results with actual rallies, ensuring that simulated rallies reflect the competitive dynamics of actual gameplay.
\end{compactitem}
Recent investigations have showcased the success of offline imitation learning, including hierarchical imitation learning (HIL) models \cite{zhang2021provable}.
Although HIL approaches enrich imitated expressivity in the long-horizontal task, none of them were designed for turn-based sports, which consist of multiple players taking actions alternatively.
Therefore, there are two challenges to applying existing HIL methods for turn-based sports:
\textbf{1) Leveraging experience.} When players encounter situations that have appeared in their experience (e.g., from training and previous matches), they usually take corresponding actions for returning shots. 
Furthermore, in the same situation, there may be a wide variety of different corresponding behaviors in their experience. 
It is challenging to leverage experience to provide the helpful information for the agent.
\textbf{2) Alternative decision-making.} In a turn-based sport, the state of each player is determined by the actions of not only themselves but also other players. 
Thus, The errors of one agent's decisions affect others, leading to more severe compounded errors than in typical multi-agent imitation learning tasks \cite{le2017data}.
\begin{figure}
    \centering
    \includegraphics[width=\textwidth]{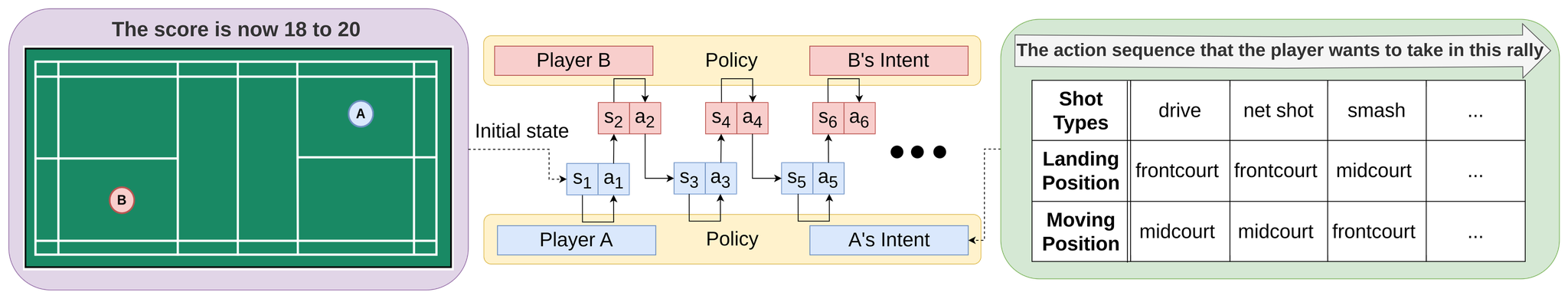}
    \caption{An illustration of RallyNet in a badminton rally.}
    \label{fig:concept}
\end{figure}

To address these challenges, we propose a hierarchical offline imitation learning model via experiential context and geometric Brownian motion (RallyNet) to capture long-term decision dependencies by modeling decision-making processes in turn-based sports as contextual Markov decision processes (CMDP) \cite{hallak2015contextual}.
Based on the CMDP setting, the \textbf{Experiential Context Selector (ECS)} was designed by establishing the context space from experiences and selecting a context in the space as the agent's intent to mimic the decision-making of the agents following their intents in the rally. This enables the agent's behavior throughout the rally to not be influenced by partially incorrect decisions and provides the physical explanations of the context to understand the intent behind the player’s behavior. 
By viewing the interactions between players as those between particles, we introduce \textbf{Latent Geometric Brownian Motion (LGBM)} to capture the interactions between players. We make players participating in a rally alternately complete geometric Brownian motion \cite{revuz2013continuous} in latent space, which enables the agent's decision-making to jointly consider the opponent's behavior. This generates more realistic behavior in alternative decision-making.
Fig. \ref{fig:concept} illustrates how RallyNet is applied to a turn-based sport, and uses a badminton singles match as an example.
An initial state of the rally including two players' positions and score information is provided to recover the content of the rally.
We highlight our main contributions as follows:
\begin{compactitem}
\item We propose a novel HIL model via experiential context and geometric Brownian motion named RallyNet to learn player decision-making strategies in turn-based sports. 
\item RallyNet treats decision-making in badminton games as CMDPs, utilizing experiences to shape agents' intent. Additionally, we introduce latent geometric Brownian motion to enhance realism by capturing player interactions.
\item We quantitatively validate the performance of RallyNet in a real-world badminton dataset with men's singles and women's singles, which demonstrates that RallyNet attains superior performance than prior offline IL methods and the state-of-the-art turn-based supervised method. surpassing them by 16\% in the mean of the rule-based agent normalized score.
\end{compactitem}

\section{Related Work}
\subsection{Inverse Reinforcement Learning}
Inverse Reinforcement Learning (IRL) \cite{ng2000algorithms} is an IL approach that attempts to learn the underlying rewards that an expert is optimizing from the demonstrations. With the learned reward function, IRL can guide the agent's behavior toward that of the expert. However, in turn-based sports, each player's actions determine the next state for other players. Treating opponents as part of the environment is equivalent to training in a constantly changing environment, which can make it challenging for IRL to learn optimal policies.

\subsection{Offline Imitation Learning via Behavior Cloning}
Behavior cloning (BC) provides a form of supervised learning for training policies by learning direct mapping from states to actions, which can be used in the offline setting \cite{pomerleau1988alvinn}. 
Recent works such as DT \cite{chen2021decision}, GCSL \cite{ghosh2019learning}, and RvS \cite{emmons2021rvs} have shown that not only is supervised learning and bypassing the learning reward function able to attain better results for offline learning, but it is also easier to use and is more stable than offline inverse reinforcement learning (IRL) \cite{ng2000algorithms,garg2021iq}. Moreover, previous work \cite{le2018hierarchical,zhang2021provable} has extended the BC to hierarchical BC by dividing a task into several sub-tasks, where the low-level behaviors are controlled by high-level decisions to capture long-term decision-making processes.
However, these existing approaches focus on the complex behavior and the long-term task of the same target, but neglect the characteristics of the mixed sequences, which causes serious compounding errors.

\section{Preliminaries}
\subsection{Badminton: A Typical Example of a Turn-Based Sport}
\label{ENV}
For simplicity, we consider a badminton game with two players (i.e., singles matches) as the demonstration example.
As shown in Fig. \ref{fig:concept}, the state of the player who hits the shuttlecock consists of the score information, the 2-dimensional position of the shuttlecock, the shot type to receive, the player's position, the opponent's position, and the moving vector of the opponent (e.g., move forward 1m).
The action of the player consists of the landing position of the shuttlecock, the shot type (the shuttlecock type to hit), and the moving position to go to after returning the shuttlecock.

\subsection{The Contextual Markov Decision Process}
\label{MDP}
A Markov Decision Process (MDP) \cite{puterman2014markov} is defined by  $(S,A,T(y|x,a),R(x),\gamma)$, where $S$ is a set of states, $A$ is a set of possible actions agents can take, $T(y|x,a)$ is the transition probability, $R(x)$ is a reward function, and $\gamma \in [0,1)$ is the discount factor.
At the $t$-th time step, the agent receives a state $s_{t} \in S$, then takes an action $a_{t} \in A$ according to a policy $\pi(a_{t}|s_{t})$.
It is noted that the reward signal and the interaction with the environment are not available in the offline setting in this paper.
The Contextual Markov Decision Process (CMDP) \cite{hallak2015contextual} leverages side information to extend the standard MDP with multiple contexts.
A CMDP is defined by a tuple $(C,S,A,M)$, where $C$ is a set of contexts and $M$ is a function which maps a context $c \in C$ to MDP parameters $M(c) = \{R^{c}, T^{c}\}$.

\subsection{Problem Formulation}
For the badminton game with two players, we denote $P_A$ as the starting player and $P_B$ as the other player for each rally.
Let $\mathcal{R} = \{T_{r}\}^{|\mathcal{R}|}_{r=1}$ denote historical rallies of matches, where the rally is composed of a sequence of state-action pairs.
The $r$-th rally is denoted as $T_{r} = \{(s_{1}^{P_A},a_{1}^{P_A}),(s_{1}^{P_B},a_{1}^{P_B}),…,(s_{|T_{r}|}^{p},a_{|T_{r}|}^{p})\}$.
At the $t$-th step, $s_{t}^{p}$ and $a_{t}^{p}$ represent the state and action of the player who takes action, respectively, where $p \in \{P_A, P_B\}$ represents that the step is related to either $P_A$ or $P_B$.
$p = P_A$ when $t=2i-1$ and $p = P_B$ when $t=2i$, where $i=1,2, \dots, |T_{r}|$.
The action consists of the landing positions, shot type, and moving positions represented by $a_{t}^{p} = \langle l_{t}^{p}, t_{t}^{p}, m_{t}^{p} \rangle$.
The imitation learning task for turn-based sports is to learn a policy that can recover players' demonstrations from a set of historical rallies $\mathcal{R}$.
Formally, for each demonstration rally in $\mathcal{R}$, given initial state $s_{1}^{a}$ of the $r$-th rally $T_{r}$, our goal is to recover the rally $T_{r}$. 

We formulate the decision-making process in badminton games as CMDP: The context space $C$ consists of contexts with the dimension $K$.
At the beginning of each rally, the agent chooses a context $c \in C$, where $c$ represents the agent's desired intent for the rally.
Subsequently, the chosen MDP corresponding to context $c$ is applied throughout the rally until termination.
A rally would terminate when any of the following termination conditions are satisfied: (i) The agent misses the shot; (ii) The shuttlecock does not come over the net; (iii) The shuttlecock falls out of the scoring area.

Moreover, we define the \textit{experience} of player $p$ with a current state $s_{t}^{p}$ as follows: For a rally that has experienced current states $s_{t}^{p}$, the experience is the action sequence of rallies $\{a_{t|p}^{p}\}$, where $t|p$ represents the step that is player $p$ taking action.
To speed up the extracting experience, we establish an experience extracting function $EXP$ to output the experience from the historical rallies corresponding to the current state. 

We create a dictionary with keys as combinations of discrete values for player positions, opponent positions, shuttlecock positions (on a 10x10 grid), and shot types, and values as action sequences from rallies. 
By discretizing these continuous variables, we can efficiently retrieve matching rallies. 
When a new state is input, the function extracts the relevant positions and shot types, retrieves the corresponding action sequences of rallies from the dictionary, and then uses these action sequences to construct the context space.

\begin{figure*}
    \centering
    \includegraphics[width=\textwidth]{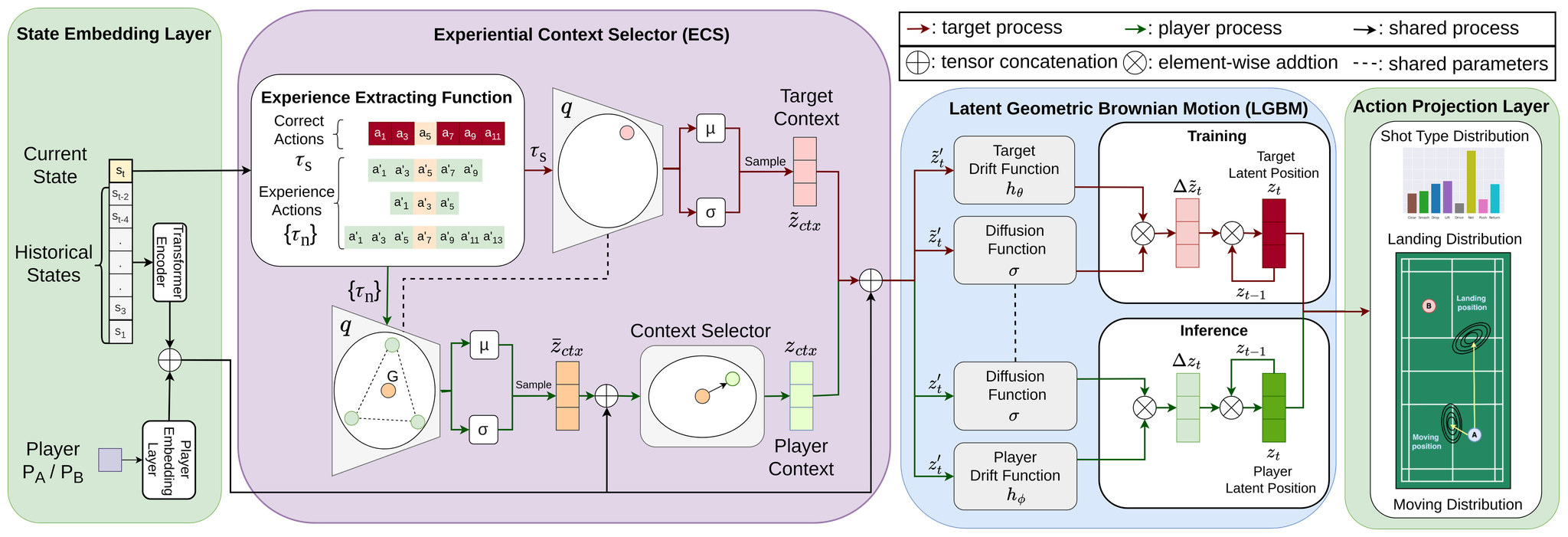}
    \caption{The framework of RallyNet.}
    \label{fig:model}
\end{figure*}

\section{Methodology}
Fig. \ref{fig:model} illustrates the RallyNet framework consisting of two components: the Experiential Context Selector (ECS) and the Latent Geometric Brownian Motion (LGBM). 
The ECS component extracts experiences and constructs a latent context space to understand player intent, ensuring the agent's behavior remains consistent during the rally.
The LGBM component frames players as particles to simulate geometric Brownian motion, effectively capturing interactions between players and allowing joint consideration of their behavior.
Notably, we designed two processes in the training stage, namely the \textit{target process} and the \textit{player process}. 
The \textit{target process} aims to learn to take actions that conform to the context by giving the correct actions as a target context for training. 
The \textit{player process} utilizes experiences to learn how to select a context that approaches the output of target processes from the context space. 
Both processes share most of the modules, and only the\textit{ player process} is used during inferencing.
We proceed to present the proposed modules. 
In this paper, for a player $p$, we use $x_{t}^{p}$ to denote the state embedding for the $t$-th step of the rally.
The state embedding is computed using a state embedding layer that processes the states experienced by the player through a Transformer encoder. The resulting embedding is then concatenated with the player embedding to obtain the final state embedding.

\subsection{Experiential Context Selector (ECS)}\label{ECS}
In turn-based sports, the high-level decisions with explicit meaning are difficult to define as the demonstrated behavior does not have an obvious hierarchical structure, and hence the conventional HIL methods are not applicable in capturing long-term decision process dependencies.
For example, in badminton games, the high-level actions typically resort to generating a rough tactical description of the entire rally like \textit{taking more crosscourt shots first and then finding a chance to block the net}.
With that said, high-level decisions are required to capture the conceptual intents, which are rather implicit and more complex than the explicit goals or sub-tasks in the conventional HIL.

To enable HIL in turn-based sports, we propose the ECS to leverage the experience and provide prior information to the agent, e.g., \textit{what context the rally will have in a current state}.
ECS leverages experience to construct a context space in which the agent is allowed to select the most relevant context as the intent of the rally, thereby ensuring that the agent's behavior during the rally is not influenced by partially incorrect decisions from the agent.
Specifically, we first extract experiences of the current state $s_{t}^{p}$ from historical rallies by the experience extracting function $EXP$:
\begin{equation}
\label{Eq:Experience Extracting}
\small
  \{\widetilde{\tau}_{t}^{p}, {{\{\tau_{n,t}^{p}\}}\} \leftarrow  EXP(s_{t}^{p}, \mathcal{R})},
\end{equation}
where $\widetilde{\tau}_{t}^{p}$ denotes the correct action sequence $\{a_{t|p}^{p}\}$, and $\{\tau_{n,t}^{p}\}$ denotes the extracted experiences.
For ease of notation, we let $\tau_{t}^{p} = \{\widetilde{\tau}_{t}^{p}, {\{\tau_{n,t}^{p}\}}\}$.

We employ the variational autoencoders (VAE) \cite{kingma2013auto,rezende2014stochastic} architecture as the context encoder and context decoder for learning meaningful context representations from experiences.
For the $t$-th step in a rally, the context encoder $q$ encodes experiences $\{\tau_{n,t}^{p}\}$ into contexts.
The context encoder $q$ encodes the $n$-th experience $\tau_{n,t}^{p}$ into context $z_{n}$:
\begin{equation}
    \label{Eq:encode_ecs}
    \small
    z_{n} \sim q(\bullet|\tau_{n,t}^{p}).
\end{equation}
ECS iteratively encodes every experience into a context and collects them as $Z_{ctx}$ for building a context space.
Following existing work (e.g., \cite{garnelo2018conditional,hamilton2017inductive}) producing latent representations, we average the collected contexts $Z_{ctx}$ to get the centroid of the context space $\bar{z}_{ctx}$:
\begin{equation}
    \label{Eq:mean_ecs}
    \small
    \bar{z}_{ctx} = mean(Z_{ctx}).
\end{equation}
We concatenate the centroid of the context space $\bar{z}_{ctx}$ and state embedding $x_{t}^{p}$ and use a linear layer as the context selector $CTX$ to select the player context $z_{ctx}$ as the agent's intent:
\begin{equation}
    \label{Eq:select_ecs}
    \small
    z_{ctx} = CTX(concat(\bar{z}_{ctx}, x_{t}^{p})).
\end{equation}

There are two processes in ECS: (i) For the $t$-th step in a rally, the player process selects the player context $z_{ctx}$ in the context space constructed based on the experiences; and
(ii) The target process utilizes the same context encoder $q$ to encode the correct action sequence of $\widetilde{\tau}_{t}^{p}$ to obtain the target context $\widetilde{z}_{ctx}$:
\begin{equation}
    \label{Eq:target_ecs}
    \small
    \widetilde{z}_{ctx} \sim q(\bullet|\widetilde{\tau}_{t}^{p}).
\end{equation}
Since the goal of the player process is to produce outputs that are close to the target process, the goal of the context selector is to select a player context that is close to the target context.

\subsection{Latent Geometric Brownian Motion (LGBM)}
\label{LGBM}
In badminton games, players consider their opponent's intent and next action to determine their defensive position accordingly to better receive the shuttlecock. As a result, a player's behavior changes depending on the opponent's intent.
To enable the agent's behavior to jointly consider the opponent's actions, LGBM is proposed to model players as particles undergoing geometric Brownian motion in latent space, which brings the inductive bias of geometric Brownian motion into the player's decision-making.
The flexibility of GBM in capturing diverse decision distributions has led to its successful application in various domains, such as finance systems \cite{finance_GBM} and population dynamics \cite{population_GBM}. These successes serve as inspiration for  capturing the inherent uncertainty and interaction in badminton.

Specifically, we first concatenated the agent's state and selected context as the agent's position in the latent space:
\begin{equation}
\small
    \begin{split}
    \widetilde{z}_{t}^{\prime} = concat(x_{t}^{p}, \widetilde{z}_{ctx}),\quad
    z_{t}^{\prime} = concat(x_{t}^{p}, z_{ctx}),\\
    \end{split}
\end{equation}
where $\widetilde{z}_{t}^{\prime}$ is the target latent position for the target process and $z_{t}^{\prime}$ is the player latent position for the player process.
Secondly, we followed the setting in \cite{li2020scalable} to simulate the geometric Brownian motion in the latent space. A standard Brownian motion is a random process and is described by the following stochastic differential equation (SDE):
\begin{equation}
\small
    dX_{t}=\sigma(x,t) dW_{t},
    \label{eq:BrM}
\end{equation}
where $W_{t}$ is Brownian motion and $\sigma$ is the diffusion function. $ \mathrm{\Delta}W_{t} \sim \sqrt{\Delta t}\,\mathcal{N}(0,1) $ is used for the simulation of the Brownian motion, where $\mathcal{N}(0,1)$ is a normal distribution with zero mean and unit variance.
The geometric Brownian motion can be generalized from Eq. (\ref{eq:BrM}) with a drift function $h(x,t)$:
\begin{equation}
\small
    dX_{t}=h(x,t)dt+\sigma(x,t) dW_{t}.
    \label{eq:recepemu}
\end{equation}
We describe the discrete-time SDE of the geometric Brownian motion in the latent space for two processes, $\mathrm{\Delta}\widetilde{z}_{t}$ and $\mathrm{\Delta}z_{t}$ as follows:
\begin{equation}
\label{Eq:SDE}
\small
\begin{split}
\mathrm{\Delta}\widetilde{z}_{t} = h_{\theta}(\widetilde{z}_{t}^{\prime}, t)\mathrm{\Delta}t + \sigma(\widetilde{z}_{t}^{\prime}, t)\mathrm{\Delta}W_{t}, \quad
\mathrm{\Delta}z_{t} = h_{\phi}(z_{t}^{\prime}, t)\mathrm{\Delta}t + \sigma(z_{t}^{\prime}, t)\mathrm{\Delta}W_{t},
\end{split}
\end{equation}
where $h_{\theta}$ and $h_{\phi}$ are the target drift function and the player drift function, respectively.
Both processes share the same function $\sigma$.
Finally, we add the displacement to the previous latent position to compute the new latent position:
\begin{equation}
\small
\begin{split}
    \label{Eq:SDE add}
    z_{t} = z_{t-1} + \mathrm{\Delta}\widetilde{z}_{t} \enspace \mbox{(during training)}, \quad
    z_{t} = z_{t-1} + \mathrm{\Delta}z_{t} \enspace \mbox{(during inference)}.
\end{split}
\end{equation}
Since the displacement can be regarded as the decision of the agent, LGBM makes the agent consider the opponent's decision through the displacement added to the previous latent position.

\subsection{Action Projection Layer}
\label{ACT}
The action projection layer is designed to project the latent position to the action that can interact with the opponent.
To predict the shot type, we apply a linear layer to the latent position $z_{t}$ to predict the shot type $\hat{t}_{t}^{p}$ at the $t$-th step:
\begin{equation}\label{Eq:action_shot}
\small
\hat{t}_{t}^{p} = softmax(W^{T}z_{t}),
\end{equation}
where $W^{T} \in \mathbb{R}^{N_{t} \times d }$ is a learnable matrix, and $N_{t}$ is the number of shot types. 
To predict the landing positions $\hat{l}_{t}^{p}$ and the moving positions $\hat{m}_{t}^{p}$, we assume the landing distribution and moving distribution are weighted bivariate normal distributions which contain the mean $\{\mu_{t}\} = \{\langle \mu_{x}, \mu_{y}\rangle_{t}\}$, standard deviation $\{\sigma_{t}\} = \{\langle\sigma_{x}, \sigma_{y}\rangle_{t}\}$, and weight $\{w_{t}\}$ of $N_{g}$ bivariate normal distributions.
We apply two linear layers to predict the parameterized distributions for the landing and moving positions:
\begin{equation}\label{Eq:action_position}
\small
\begin{split}
\langle \{\mu_{t}^{L}\}, \{\sigma_{t}^{L}\}, \{w_{t}^{L}\} \rangle = W^{L}z_{t}, \quad
\langle \{\mu_{t}^{M}\}, \{\sigma_{t}^{M}\}, \{w_{t}^{M}\} \rangle = W^{M}z_{t},
\end{split}
\end{equation}
where $W^{L} \in \mathbb{R}^{5 \times N_{g} \times d}$ and $W^{M} \in \mathbb{R}^{5 \times N_{g} \times d }$ are two learnable matrices for landing and moving, respectively. 
Finally, we concatenate the shot types and both landing and moving positions to get the action of agent $\hat{a}_{t}^{p} = \langle \hat{t}_{t}^{p}, \hat{l}_{t}^{p}, \hat{m}_{t}^{p} \rangle$ and decide the next state of the opponent.

\subsection{Loss Function}
To mimic the player’s action at each step, we minimize the loss as:
\begin{equation}
\small
\mathcal{L}  =  w_{1} \cdot \mathcal{L}_{pred} + w_{2} \cdot \mathcal{L}_{ctx} + w_{3} \cdot \mathcal{L}_{sde},
\end{equation}
where $w_{1}$, $w_{2}$, $w_{3}$ $\in$ [0, 1] are hyper-parameters to balance the weights of the corresponding losses. 
\begin{equation}
\small
\begin{split}
\mathcal{L}_{pred} = \mathcal{L}_{type} + \mathcal{L}_{land} + \mathcal{L}_{move} + \mathcal{L}_{reg}, \quad
\mathcal{L}_{type} = - \sum_{T_r \in \{\mathcal{R}\}} \sum_{t=1}^T   t_{t}^{p}\log \hat{t}_{t}^{p},
\end{split}
\end{equation}
where $\mathcal{L}_{type}$ is the cross-entropy loss for the predicted shot types.
$\mathcal{L}_{land}$ and $\mathcal{L}_{move}$ are the negative log-likelihood losses for the prediction of both landing and moving positions. 
To simplify the expression, we use $\mathcal{L}_{(land,\ move)}$ to represent $\mathcal{L}_{land}$ or $\mathcal{L}_{move}$:
\begin{equation}
\small
\begin{split}
\mathcal{L}_{(land,\ move)} = - \sum_{T_r \in \{\mathcal{R}\}} \sum_{t=1}^T  
\log(\mathcal{P}(x_{t}^{(L,\ M)}, y_{t}^{(L,\ M)}| \{\mu_{t}^{(L,\ M)}\}, \{\sigma_{t}^{(L,\ M)}\}, \{w_{t}^{(L,\ M)}\})).
\end{split}
\end{equation}
The regularization loss $\mathcal{L}_{reg}$ is introduced to prevent the model from degenerating into a simple strategy by ensuring the avoidance of overlapping distributions, computed as the average negative distance between their means.

The context encoder encodes each experience and generates the mean and standard deviation for each context. These values are used to sample the context embedding, which is then passed to the context decoder for experience reconstruction.
Therefore, $\mathcal{L}_{ctx}$ consists of the latent loss $\mathcal{L}_{latent}$ and the reconstruction loss $\mathcal{L}_{recon}$.
The latent loss $\mathcal{L}_{latent}$ is the KL divergence between the context space distribution and the standard Gaussian distribution $\mathcal{N}(0,1)$ (with zero mean and unit variance). 
Specifically, $\mathcal{L}_{latent}$ can be expressed as:
\begin{equation}
\small
\mathcal{L}_{latent} = D_{KL}(\mathcal{N}(\mu_{c}, \sigma_{c}) || \mathcal{N}(0,1)),
\end{equation}
where $\mu_{c}$ and $\sigma_{c}$ are the mean and standard deviation output by the context encoder, respectively.
The objective of context decoders is to reconstruct intent, which are action sequences of rallies, from selected contexts.
Therefore, the reconstruction loss $\mathcal{L}_{recon}$ is the same as $\mathcal{L}_{pred}$.
Finally, we follow \cite{li2020scalable} to minimize the KL divergence $\mathcal{L}_{sde}$ between two SDEs in Eq. (\ref{Eq:SDE}) of LGBM:
\begin{equation}
\small
\mathcal{L}_{sde} = \sum_{T_r \in \{\mathcal{R}\}} \sum_{t=1}^T  \frac{1}{2} \lvert (h_{\phi}(\widetilde{z}_{t}^{\prime}, t) - h_{\theta}(\widetilde{z}_{t}^{\prime}, t)) / \sigma(\widetilde{z}_{t}^{\prime}, t) \rvert^{2}.
\end{equation}

\section{Experiments}
To evaluate RallyNet's imitation performance, we conducted comprehensive experiments addressing three key criteria and explored case studies to gauge its value to the badminton community\footnote{Our code and dataset are already publicly available as the GitHub repository \href{https://github.com/wywyWang/CoachAI-Projects}{https://github.com/wywyWang/CoachAI-Projects}.}:

\begin{compactitem}
    \item [RQ1] \textbf{Behavioral Sequence Similarity:} Do the sequences of shot types, shuttlecock landing positions, and player moving positions in rallies generated by RallyNet closely match those in real rallies?
    \item [RQ2] \textbf{Rally Duration Realism:} Does RallyNet replicate the length of rallies, providing a realistic representation of the duration observed in real matches?
    \item [RQ3] \textbf{Outcome Consistency:} Are the outcomes generated by RallyNet consistent with the real-world results?
    Specifically, does the win rate of rallies simulated by the learned agent align with the win rate of real players?
    \item [RQ4] \textbf{Case Studies:} How does RallyNet's capability to replicate player behavior bring valuable applications to the badminton community?
\end{compactitem}

\subsection{Experimental Setup}
\noindent\textbf{Badminton Dataset.}
RallyNet is evaluated on the largest badminton singles dataset, which is the only publicly available benchmark in turn-based sports \cite{wang2022shuttlenet,DBLP:conf/aaai/ChangWP23}.
This dataset comprises 75 singles matches played by 31 players from 2018 to 2021, totaling 180 sets, 4,325 rallies, and 43,191 strokes.
To construct the action space, we used the 12 shot types defined by \cite{10.1145/3551391}, namely \textit{receiving}, \textit{short service}, \textit{long service}, \textit{net shot}, \textit{clear}, \textit{push/rush}, \textit{smash}, \textit{defensive shot}, \textit{drive}, \textit{lob,} \textit{drop}, and \textit{can't reach}.
For each match, the 80\% of rallies are designated as training data, while the remaining 20\% are reserved for testing, ensuring comprehensive integration of each player's historical data into the model. We utilize 5-fold cross-validation to fine-tune hyperparameters.

\noindent\textbf{Baselines.} 
We compare RallyNet against several baselines, including: 
1) \textbf{Random agent}, which samples the shot type as well as both the landing and moving positions uniformly randomly; 
2) \textbf{Rule-based agent}, which samples from the distribution of the shot type and from both landing and moving distributions which are obtained through the experience extracting function; 
3) \textbf{IQ-Learn} \cite{garg2021iq}, a state-of-the-art model-free offline IRL; 
4) \textbf{Behavior Cloning (BC)} \cite{pomerleau1988alvinn}; 
5) \textbf{Hierarchical Behavioral Cloning (HBC) }\cite{zhang2021provable}, which learns an options-type hierarchical policy from demonstrations; 
6) \textbf{ShuttleNet} \cite{wang2022shuttlenet}, which is a specialized sequence-to-sequence model designed for stroke forecasting. 
It fuses the contexts of rally progress and player styles to predict the shot type and the landing position based on past information. 
It is noted that ShuttleNet is not designed for imitation; requiring at least two steps to encode players' contexts, it cannot be tested solely from the initial state.
7) \textbf{DyMF} \cite{DBLP:conf/aaai/ChangWP23}, which is a specialized sequence-to-sequence model tailored for badminton movement forecasting. 
It employs dynamic graphs and hierarchical fusion to adeptly capture player interactions. 
Similar to ShuttleNet, DyMF relies on past information for future predictions, limiting its ability to recover the entire rally from an initial state.

\noindent\textbf{Evaluation Metrics.} 
Given the absence of prior research on the imitation of behaviors in turn-based sports, we proposed 4 evaluation metrics to measure the similarity between generated rallies and true player rallies.
To evaluate the results of shot type prediction, we used Connectionist Temporal Classification (CTC) loss \cite{graves2006connectionist} for uncertainty measurement, which is defined as the negative log-likelihood of the labels given input sequences.
To evaluate the results of the predicted landing and moving positions, we used Dynamic Time Warping (DTW) \cite{berndt1994using} to calculate the distance between generated position sequence and the true position sequence, which can assess the similarity of two sequences at a global level, as used recently in \cite{song2021ag}.
Using the DTW distance and the CTC loss has the benefit that the similarity of two sequences can be assessed even if each sequence is of a different length.
Moreover, inspired by \cite{badia2020agent57}, we further provide an overall comparison of the algorithms by introducing a metric termed Rule-based agent Normalized Score (RNS) defined as: 
$
RNS = \frac{(Random_{score}-Agent_{score})}{(Random_{score}-Rule_{score})}, 
$
where the subscript $score$ indicates the metric used and can be either DTW distance or CTC loss, and $Agent_{score}$ is the performance of the agent. $Random_{score}$ and $Rule_{score}$ are the performance of the random agent and the rule-based agent, respectively.
To provide an overall comparison, we present the Mean RNS (MRNS), which is defined as the average over the RNS scores under the 3 metrics.
This metrics aligns with our emphasis on tactical execution, where Mean Squared Error (MSE) or Cross-Entropy (CE) used in \cite{wang2022shuttlenet,DBLP:conf/aaai/ChangWP23} might overlook similarities within decision sequences. For instance, if a strategy compels opponents to traverse the court substantially, a minor delay by a single stroke should still be recognized as similar.
All the results are the average of 5 different random seeds.

\begin{table*}[t]
\centering
\caption[c]{Quantitative results. The best is in boldface and the second best is underlined. Since ShuttleNet and DyMF predict based on past information, the symbol $-$ denotes that the result is unavailable.}
\scalebox{0.92}{
    \begin{tabular}{c|c c c c|c c c c}
    \toprule
    &\multicolumn{4}{c|}{Given initial state only} &\multicolumn{4}{c}{Given states of the first two steps} \\
    \cmidrule{2-9}
    Model & Land($\downarrow$) & Shot($\downarrow$) & Move($\downarrow$) & MRNS($\uparrow$) 
    & Land($\downarrow$) & Shot($\downarrow$) & Move($\downarrow$) & MRNS($\uparrow$) \\
    \midrule
    Random agent & 1.3044 & 104.382 & 0.8506 & - & 1.2548 & 79.2586 & 0.8763 & - \\
    Rule-based agent & 0.9130 & 65.4033 & 0.5942 & 1 & 1.0888 & 61.7431 & 0.6464 & 1 \\
    IQ-Learn \cite{garg2021iq} & 1.7464 & 127.8784 & 1.072 & -0.8652 & 1.6454 & 112.0538 & 1.0555 & -1.6682 \\
    BC \cite{pomerleau1988alvinn} & 0.9603 & 63.5967 & 0.4829 & 1.1199 & 1.0343 & 56.1007 & 0.5832 & 1.3084\\
    HBC \cite{zhang2021provable}& \underline{0.6940} & \underline{32.5972} & \underline{0.4031} & \underline{1.7155} & \underline{0.8760} & 33.1886 & 0.5529 & \underline{2.1062}\\
    DyMF \cite{DBLP:conf/aaai/ChangWP23} & - & - & - & - & 1.1828 & \underline{26.5931} & 1.2266 & 0.6389 \\
    ShuttleNet \cite{wang2022shuttlenet} & - & - & - & - & 0.9090 & 34.0467 & \underline{0.5059} & 2.0918 \\
    \midrule
    RallyNet (Ours) & \textbf{0.5931} & \textbf{18.8678} & \textbf{0.3416} & \textbf{1.9988} & \textbf{0.7959} & \textbf{19.5100} & \textbf{0.4943} & \textbf{2.6124}\\
    \bottomrule
    \end{tabular}
}
\label{tab:task}
\end{table*}

\subsection{Quantitative Results (RQ1)}
Table \ref{tab:task} presents the performance of RallyNet and the baselines. 
We summarize the observations as follows: 

\noindent\textbf{Superior Performance of RallyNet.} 
RallyNet surpasses all the baselines in terms of all metrics, whether given only the initial state, or the state of the first two steps. Quantitatively, RallyNet outperforms all the baselines by at least 7.08 in terms of CTC loss when predicting the shot types, and achieves lower DTW distances than the baselines by at least 8.27\% and 1.16\% in the predicted landing and moving positions, respectively. 
Note that since players usually have similar positions and shot types for \textit{serving} and \textit{receiving} (i.e., the first two steps), the prediction after two steps is more difficult. 
These findings highlight RallyNet's effective mitigation of compounding errors in badminton player imitation.

\noindent\textbf{Hierarchical BC Advantage.}
We observe that IQ-Learn produces unfavorable performance, due to the difficulty of inferring the reward function in a scenario where the opponent exclusively determines the next state. 
This may lead to a weak correlation between the current and the next state, making this IRL method unsuitable for this particular task.
HBC can better capture complex behavior as it can switch between different strategies by selecting different options based on varying states, highlighting the importance of using HIL in turn-based sports.

\noindent\textbf{The Importance of Leveraging Experience.} 
Existing supervised learning methods for turn-based sports, such as ShuttleNet and DyMF, diverge from imitation learning models. 
These models, structured on sequence-to-sequence architectures, predict player actions by relying on historical rally data. 
DyMF excels in predicting shot types, while ShuttleNet performs well in forecasting player movements, showcasing its proficiency in the handling of player interactions and rally dynamics. However, these approaches face challenges as errors accumulate during prolonged rallies.
In contrast, RallyNet stands out by leveraging experience, enabling it to emulate player-like behavior even in long rallies. 

\subsection{Length Distribution Difference (RQ2)}
\begin{figure}
    \centering
    \includegraphics[width=0.85\textwidth, height=4.25cm]{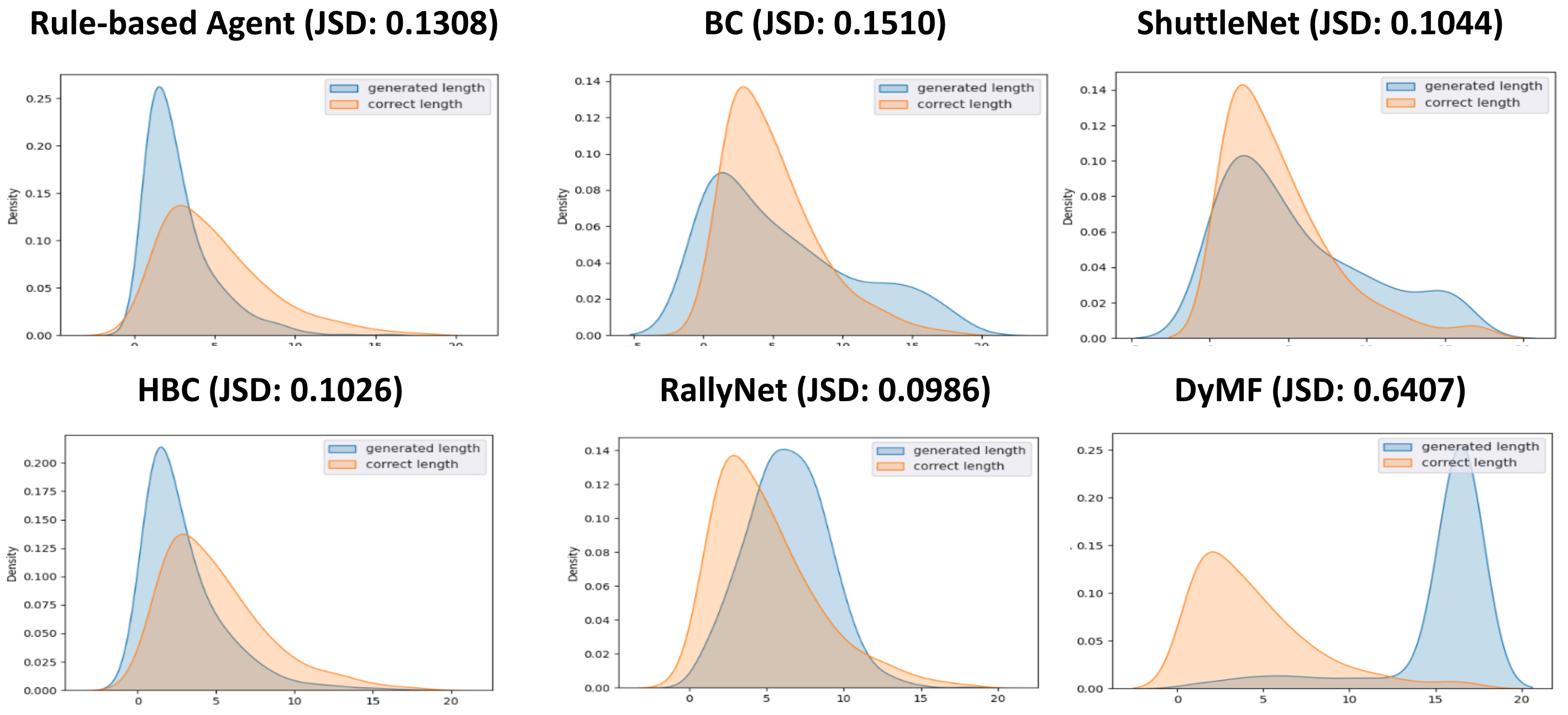}
    \caption{The length distribution difference. The orange distribution represents the ground truth rally length distribution, while the blue distribution depicts the generated rally length distribution.}
    \label{fig:ldd}
\end{figure}

The length of generated rallies serves as a reflective measure of whether the model has learned realistic gameplay situations. 
To visually depict the distribution of generated rally lengths in comparison to real rally lengths, we utilize Kernel Density Estimate (KDE). 
Additionally, the quantitative dissimilarity between these two distributions is evaluated using the Jensen–Shannon divergence (JSD).
In Fig. \ref{fig:ldd}, BC struggles to effectively capture player interactions, resulting in premature or delayed termination. 
In contrast, both HBC and RallyNet, by modeling more intricate strategies, produce generated rally length distributions close to the real ones. 
Notably, even a rule-based agent using our carefully designed experience extraction function can achieve a similar rally length distribution, emphasizing that our experience extraction captures rally duration to aid RallyNet in making more realistic decisions.
However, models like ShuttleNet and DyMF, being sequence-to-sequence models rather than imitation learning models, face challenges in generating ideal rally lengths.
This challenge stems from the infrequent occurrence of error behaviors in dataset, where a rally involves only one error behavior (e.g., shuttlecock out of bounds or hitting the net)
, making it difficult for these models to accurately capture rally duration.

\subsection{Win Rate Difference (RQ3)}
To evaluate the outcome consistency of learned models in mimicking real player behavior, we adopt a straightforward approach by examining win rates among players. 
As depicted in Table \ref{tab:rollout}, considering the diverse quantities of match data available for each player in the dataset, our analysis focuses on two real-world professional players, X and Y, who possess the most extensive match records. For both X and Y, we chose four opponents from their respective match histories, subsequently comparing the generated rally win rates with the ground truth win rates. 
To ensure the validity of win rate observations within realistic gameplay scenarios, we evaluated the performance of the top three imitation learning models: BC, HBC, and RallyNet. 
Table \ref{tab:rollout} illustrates that RallyNet consistently demonstrates smaller differences from the actual win rate across a majority of matchups, with a maximum difference of merely 8.15\% and an average difference of 3.79\%. In contrast, BC and HBC exhibit differences of up to 22.5\% in specific matchup combinations, with an average difference surpassing RallyNet by at least 7.16\%. 
These findings not only underscore RallyNet's stability but also highlight its exceptional capability to accurately replicate real match outcomes.

\begin{table*}
    \centering
    \caption[c]{Win rate differences between ground truths and learned agents. X and Y denote two actual players chosen from the dataset. A, B, C, D are the four opponents of player X, while E, F, G, H are the four opponents of player Y. Lower differences indicate better alignment between observed and predicted win rates by the agents.}
    \scalebox{0.80}{
        \begin{tabular}{c|c c c c|c c c c | c} 
        \toprule
        &\multicolumn{4}{c|}{X} &\multicolumn{4}{c|}{Y}\\
       \cmidrule{2-9}
        Opponent & A & B & C & D & E & F & G & H & Mean Difference\\
        \midrule
    Ground Truth Win Rate & 0.4406 & 0.5760 & 0.6250 & 0.5945 & 0.5416 & 0.4000 & 0.5454 & 0.4615 & - \\
    \midrule
    BC & 0.1413 & \underline{0.0508} & 0.1042 & \underline{0.0810} & 0.2084 & \underline{0.2000} & 0.1364 & \underline{0.1923} & 0.1393 \\
    HBC & \underline{0.0806} & 0.1316 & \underline{0.0773} & \textbf{0.0493} & \textbf{0.0153} & 0.2250 & \underline{0.0749} & 0.2227 & \underline{0.1095}\\
    RallyNet (Ours) & \textbf{0.0760} & \textbf{0.0000} & \textbf{0.0208} & \underline{0.0810} & \underline{0.0416} & \textbf{0.0000} & \textbf{0.0455} & \textbf{0.0385} & \textbf{0.0379}\\
        \bottomrule
        \end{tabular}
    }
    \label{tab:rollout}
\end{table*}

\subsection{Case Studies (RQ4)}
\begin{figure}
\centering
\begin{minipage}[t]{0.56\textwidth}
\centering
\includegraphics[width=\textwidth]{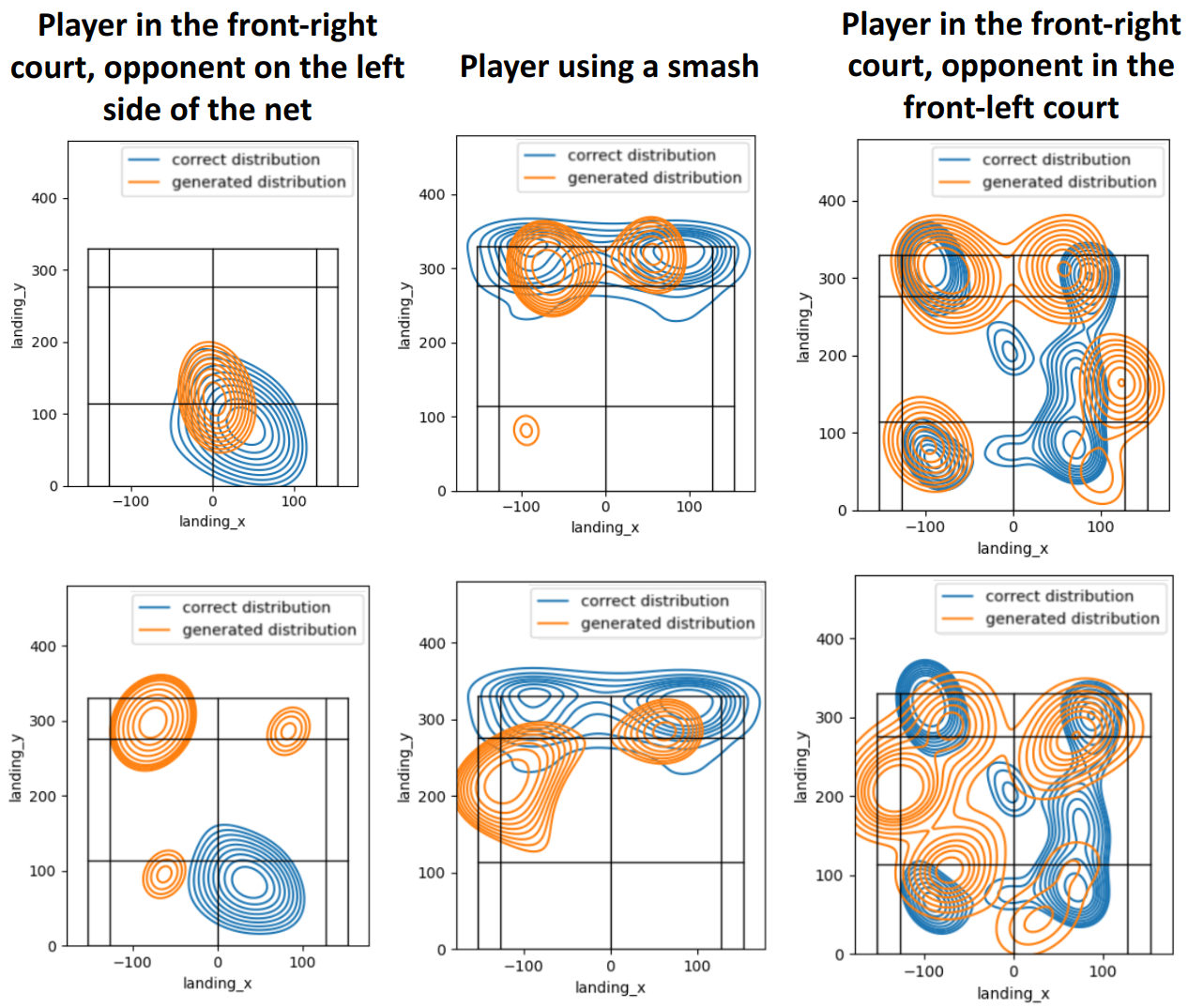}
\caption{The landing distributions of the player in various states are shown, with RallyNet in the upper row and BC in the lower row.}\label{fig:case3}
\end{minipage}
\hspace{1pt}
\begin{minipage}[t]{0.42\textwidth}
\centering
\includegraphics[width=\textwidth]{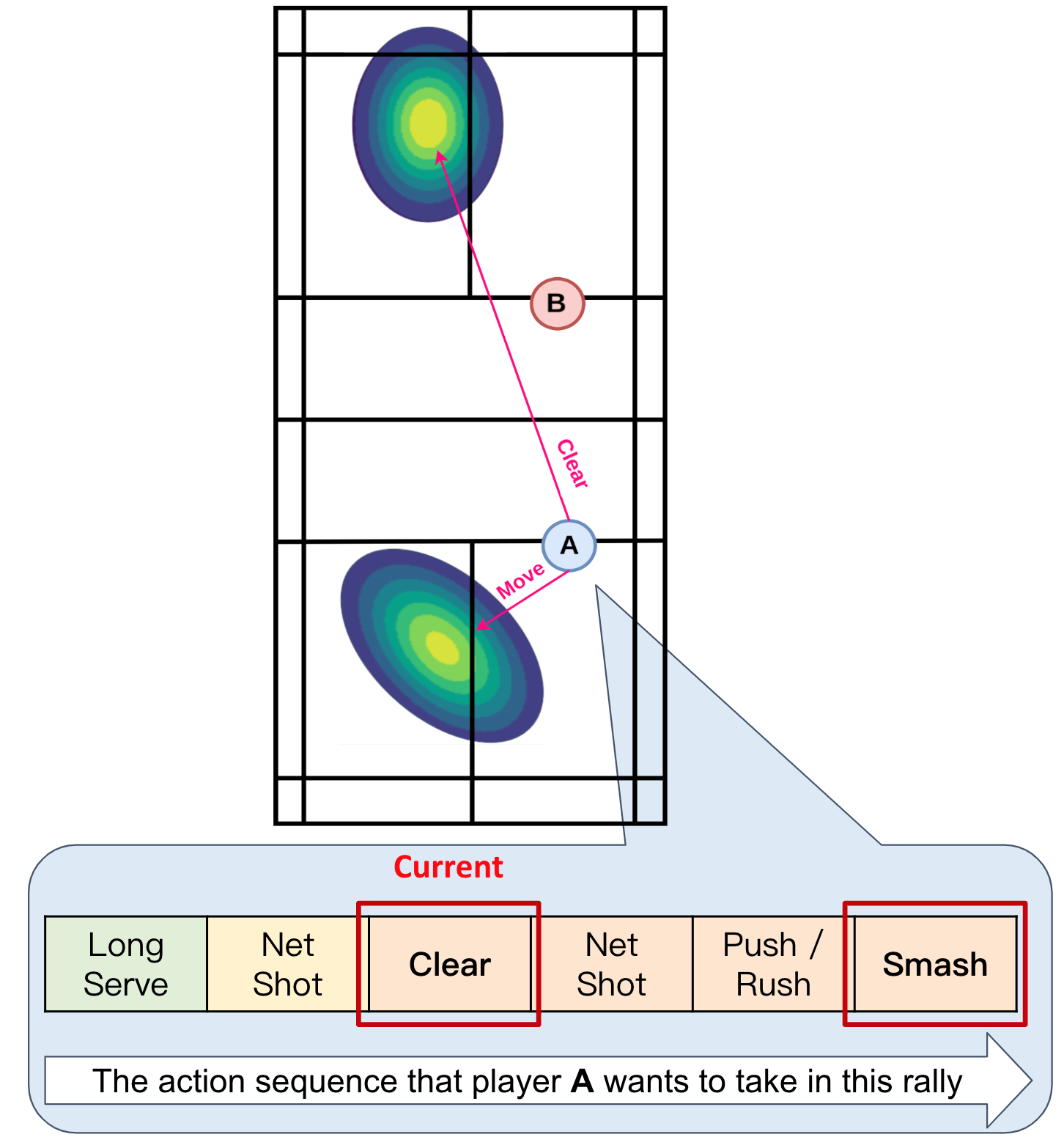}
\caption{Interpreting the intent behind an action. player A's \textit{clear} stroke is meant to plan for a \textit{smash}.}\label{fig:case2}
\end{minipage}
\end{figure}

\noindent\textbf{Simulation of Player Behavior.}
We describe a use case of the learned model in sports analytics for characterizing the style of a professional player by simulating the behavior of the player in different conditions.
We generated visualizations comparing the landing distributions of RallyNet and BC across various states. 
Fig. \ref{fig:case3} illustrates the simulated landing distributions of under different state conditions. It demonstrates that RallyNet consistently captures players' decision patterns with precision. 
In contrast, BC exhibits instances where it approximates the distribution roughly but with large deviations, and occasionally even exhibits significant misjudgments.
Accordingly, such characterization can help the coach understand the style of the player and thereby devise tactics.

\noindent\textbf{Tactical Interpretation of Player Behavior.} 
RallyNet strategically interprets player behavior by selecting a context as the agent’s intent, guiding its actions accordingly. The intent can be decoded by the context decoder of the ECS, revealing the agent's expected action sequence.
Fig. \ref{fig:case2} illustrates an example to investigate the underlying intent behind a shot based on the previous shot (long serve in the green grid) and the current shot (net shot in the yellow grid). 
For instance, if Player A's intent is decoded as planning a subsequent \textit{smash} by strategically positioning the opponent in the backcourt, it suggests a tactical move to set up aggressive shot returns. 
This showcases RallyNet's capability to provide insightful tactical interpretations, a valuable asset for badminton coaches and players.

\section{Conclusion}
In this work, we present RallyNet, a novel hierarchical offline imitation learning model for learning player decision-making strategies in turn-based sports.
By modeling players' decision-making processes as CMDP, our ECS leverages experiences to construct a context space and selects the context as the agent's intent, allowing the reduction of decision-making errors in agents' overall behaviors.
Meanwhile, LGBM captures player interactions to generate more realistic behaviors.
Extensive evaluations on real-world badminton singles matches show that RallyNet outperforms existing offline IL as well as the state-of-the-art turn-based supervised method with multifactorial assessments, and showcases the practicability for sports analytics.
We believe RallyNet serves as a general framework for other turn-based sports due to the flexible design for modeling player intents and offering an understanding of real interactions.

\end{document}